\documentclass[10pt,twocolumn,letterpaper]{article}

\usepackage{cvpr}              

\usepackage{graphicx}
\usepackage{amsmath}
\usepackage{amssymb}
\usepackage{booktabs}
\usepackage{multirow}
\usepackage{bbding}
\usepackage{algorithm}
\usepackage{algorithmic}
\usepackage{bm}

\usepackage[pagebackref,breaklinks,colorlinks]{hyperref}

\usepackage[capitalize]{cleveref}
\crefname{section}{Sec.}{Secs.}
\Crefname{section}{Section}{Sections}
\Crefname{table}{Table}{Tables}
\crefname{table}{Tab.}{Tabs.}

\def\cvprPaperID{1781} 
\def\confName{CVPR}
\def\confYear{2023}

\begin{document}

\title{T-SEA: Transfer-based Self-Ensemble Attack on Object Detection}

\author{Hao Huang\footnotemark[1]\\
Peking University\\
Beijing, China\\
{\tt\small huanghao@stu.pku.edu.cn}
\and
Ziyan Chen\footnotemark[1]\\
Peking University\\
Beijing, China\\
{\tt\small chen.ziyan@outlook.com}
\and
Huanran Chen\footnotemark[1]\\
Peking University\\
Beijing, China\\
{\tt\small huanran\_chen@outlook.com}
\and
Yongtao Wang\footnotemark[2]\\
Peking University\\
Beijing, China\\
{\tt\small wyt@pku.edu.cn}
\and
Kevin Zhang\\
Peking University\\
Beijing, China\\
{\tt\small kevinzyz@pku.edu.cn}
}
\maketitle

\renewcommand{\thefootnote}{\fnsymbol{footnote}} 
\footnotetext[1]{These authors contributed equally to this work.} 
\footnotetext[2]{Corresponding author.}

\begin{abstract}

Compared to query-based black-box attacks, transfer-based black-box attacks do not require any information of the attacked models, which ensures their secrecy. However, most existing transfer-based approaches rely on ensembling multiple models to boost the attack transferability, which is time- and resource-intensive, not to mention the difficulty of obtaining diverse models on the same task. To address this limitation, in this work, we focus on the single-model transfer-based black-box attack on object detection, utilizing only one model to achieve a high-transferability adversarial attack on multiple black-box detectors. Specifically, we first make observations on the patch optimization process of the existing method and propose an enhanced attack framework by slightly adjusting its training strategies. Then, we analogize patch optimization with regular model optimization, proposing a series of self-ensemble approaches on the input data, the attacked model, and the adversarial patch to efficiently make use of the limited information and prevent the patch from overfitting. The experimental results show that the proposed framework can be applied with multiple classical base attack methods (e.g., PGD and MIM) to greatly improve the black-box transferability of the well-optimized patch on multiple mainstream detectors, meanwhile boosting white-box performance. Our code is available at \url{https://github.com/VDIGPKU/T-SEA}.
   
\end{abstract}
\vspace{-5mm}
\section{Introduction}
\label{sec:introduction}

With the rapid development of computer vision, deep learning-based object detectors are being widely applied to many aspects of our lives, many of which are highly related to our personal safety, including autonomous driving and intelligent security. Unfortunately, recent works~\cite{huang2021rpattack,xie2017adversarial, li2018robust, wei2019transferable} have proved that adversarial examples can successfully disrupt the detectors in both digital and physical domains, posing a great threat to detector-based applications. Hence, the mechanism of adversarial examples on detectors should be further explored to help us improve the robustness of detector-based AI applications.

\begin{figure}[t]
  \centering
  \includegraphics[width=8cm]{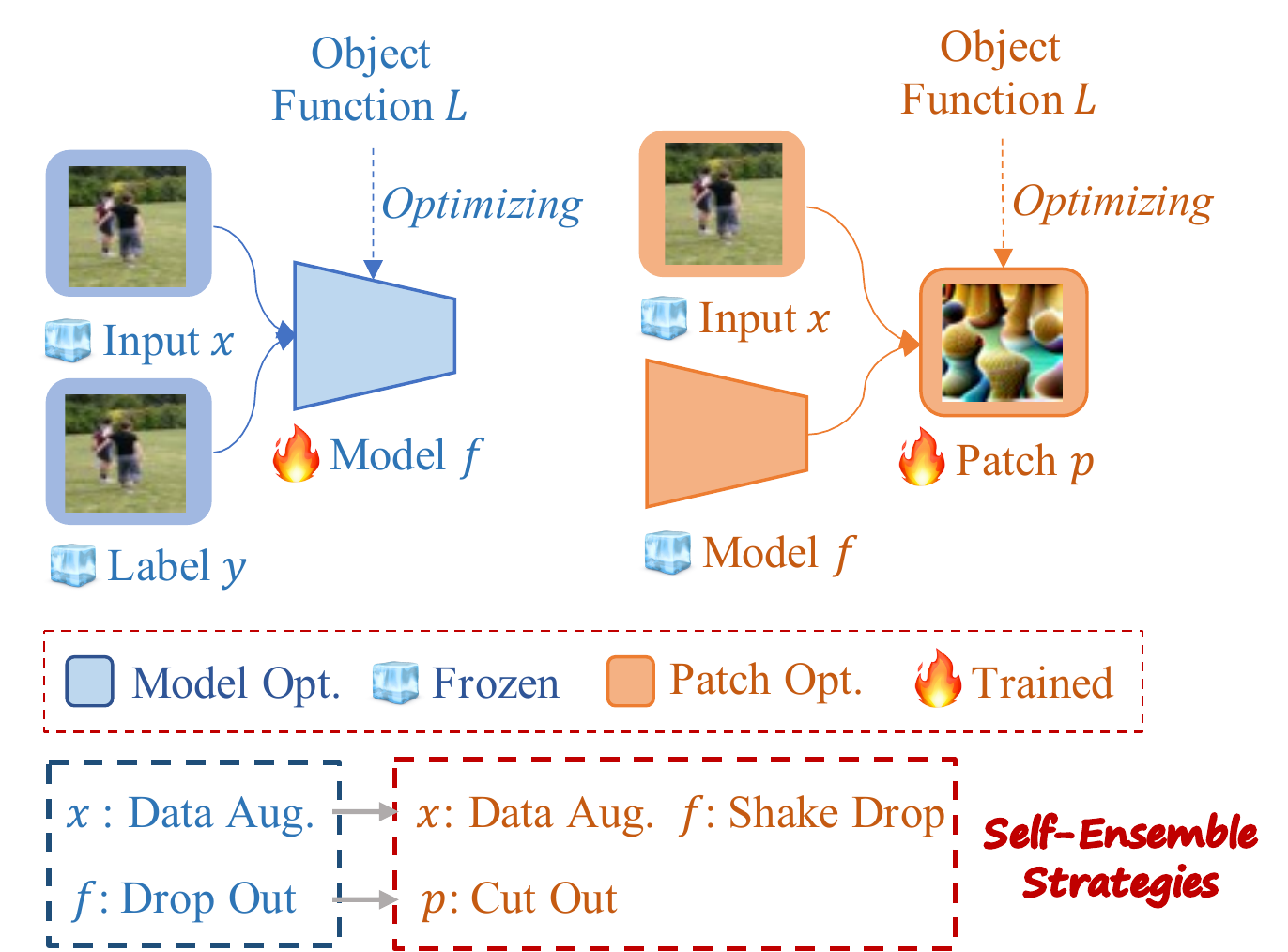}
  \caption{Model optimization usually augments the training data and drop out neurons to increase generalization, motivating us to propose self-ensemble methods for adversarial patch optimization. Specifically, inspired by data augmentation in model optimization, we augment the data $x$ and the model $f$ via constrained data augmentation and model ShakeDrop, respectively. Meanwhile, inspired by drop out in model optimization, we cut out the training patch $\tau$ to prevent it overfitting on specific models or images.}
  \vspace{-6mm}
  \label{fig:demo}
\end{figure}

In real scenes, attackers usually cannot obtain the details of the attacked model, so black-box attacks naturally receive more attention from both academia and industry. Generally speaking, black-box adversarial attacks can be classified into 1) query-based and 2) transfer-based. For the former, we usually pre-train an adversarial perturbation on the white-box model and then fine-tune it via the information from the target black-box model, assuming we can access the target model for free. However, frequent queries may expose the attack intent, weakening the covertness of the attack. Contrarily, transfer-based black-box attacks utilize the adversarial examples' model-level transferability to attack the target model without querying and ensure the secrecy of the attack. Thus, how to enhance the model-level transferability is a key problem of transfer-based black-box attacks. Most existing works apply model ensemble strategies to enhance the transferability among black-box models, however, finding proper models for the same task is not easy and training adversarial patch on multiple models is laborious and costly. To address these issues, in this work, we focus on how to enhance the model-level transferability with only one accessible model instead of model ensembling.

Though the investigation of adversarial transferability is still in its early stage, the generalizability of neural networks has been investigated for a long time. Intuitively, the association between model optimization and patch optimization can be established by shifting of formal definitions. Given the input pair of data $x\in \mathcal{X}$, and label $y\in\mathcal{Y}$, the classical model learning is to find a parametric model $f_\theta$ such that $f_\theta(x)=y$, while learning an adversarial patch treats the model $f_\theta$, the original optimization target, as a fixed input to find a hypothesis $h$ of parametric patch $\tau$ to corrupt the trained model such that $h_\tau(f_\theta, x)\neq y$. Hence, it is straight forward to analogize patch optimization with regular model optimization. Motivated by the classical approaches for increasing model generalization, we propose our \textbf{T}ransfer-based \textbf{S}elf-\textbf{E}nsemble \textbf{A}ttack (\textbf{T-SEA}), ensembling the input $x$, the attacked model $f_\theta$, and the adversarial patch $\tau$ from \textbf{themselves} to boost the adversarial transferability of the attack.

Specifically, we first introduce an enhanced attack baseline based on \cite{thys2019fooling}. Observing from Fig.~\ref{fig:scheduler} that the original training strategies have some limitations, we slightly adjust its learning rate scheduler and training patch scale to revise \cite{thys2019fooling} as our enhanced baseline (E-baseline). Then, as shown in Fig.~\ref{fig:demo}, motivated by \textit{input augmentation} in model optimization (\eg, training data augmentation), we introduce the constrained data augmentation (data self-ensemble) and model ShakeDrop (model self-ensemble), virtually expanding the inputs of patch optimization (i.e., the input data $x$ and the attacked model $f$) to increase the transferability of the patch against different data and models. Meanwhile, motivated by the \textit{dropout} technique in model optimization, which utilizes sub-networks of the optimizing model to overcome overfitting and thus increasing model generalization, we propose patch cutout (patch self-ensemble), randomly performing cutout on the training patch $\tau$ to overcome overfitting. Through comprehensive experiments, we prove that the proposed E-baseline and self-ensemble strategies perform very well on widely-used detectors with mainstream base attack methods (\eg, PGD~\cite{madry2017towards}, MIM~\cite{dong2018boosting}). Our contributions can be summarized as the following:

\begin{itemize}
    \item We propose a transfer-based black-box attack T-SEA, requiring only one attacked model to achieve a high adversarial transferability attack on object detectors.
    
    \item Observing the issues of the existing approach, we slightly adjust the training strategies to craft an enhanced baseline and increase its performance.
    
    \item Motivated by approaches increasing generalization of deep learning model, we propose a series of strategies to self-ensemble the input data, attacked model, and adversarial patch, which significantly increases the model-level adversarial transferability without introducing extra information.
    
    \item The experimental results demonstrate that the proposed T-SEA can greatly reduce the mAP on multiple widely-used detectors on the black-box setting compared to the previous methods, while concurrently performing well with multiple base attack methods.
\end{itemize}

\section{Related Work}
\label{sec:related_work}

\begin{figure*}[t]
  \centering
  \includegraphics[width=17cm]{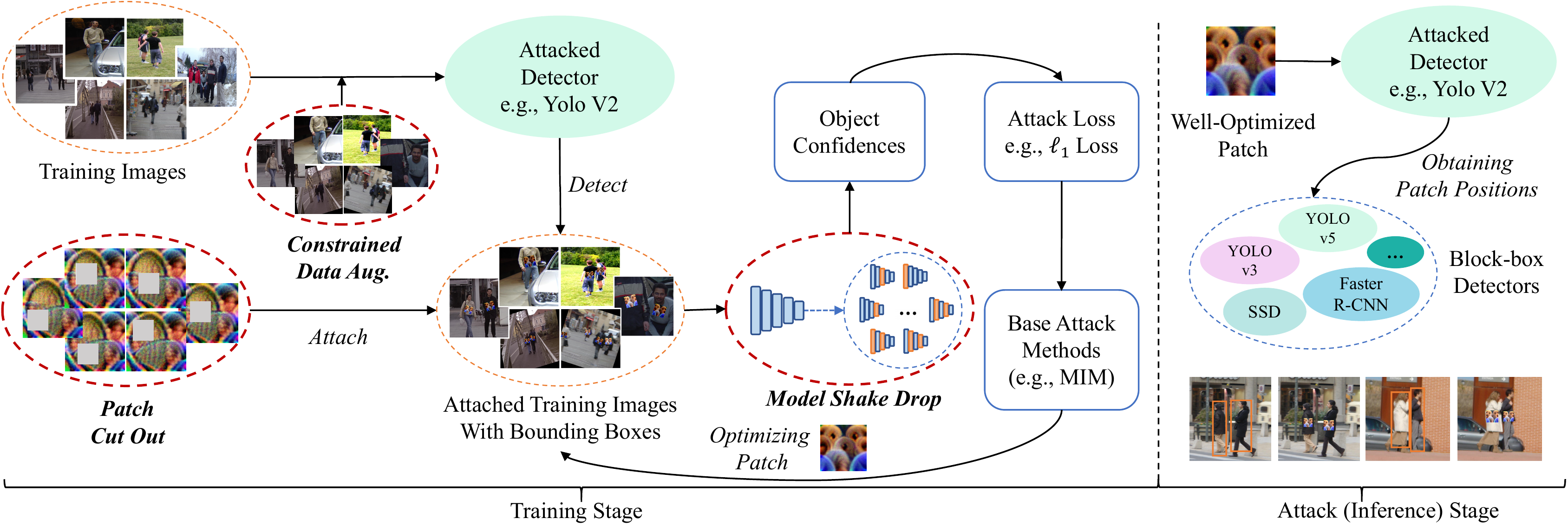}
  \vspace{-2mm}
  \caption{The Overall Pipeline of T-SEA. During the training stage, we utilize the self-ensemble strategies (i.e., the constrained data augmentation, patch cutout and model ShakeDrop) to enhance the transferability of the well-optimized adversarial patch. During the attack (inference) stage, we attach the crafted patch into different images to disrupt the detection process of multiple widely-used detectors on black-box setting.}
  \vspace{-5mm}
  \label{fig:pipeline}
\end{figure*}

\subsection{Black-box Adversarial Attack}

Since the discovery of adversarial attack by \cite{szegedy2014intriguing}, black-box attacks have gradually attracted more attention, owing to the difficulty of obtaining details of the attacked model in real scenes. Black-box attack can be separated into query-based methods~\cite{liu2022efficient, brendel2017decision, dong2020benchmarking} and transfer-based methods~\cite{papernot2017practical, bai2020adversarial}. The former utilizes the outputs of the target model to optimize the adversarial examples, which may disclose the attack behavior due to the frequent queries. Hence, in our work, we concentrate on the latter, optimizing adversarial examples in a substitute model without relying on any knowledge of the black-box models. Though the transfer-based attacks ensure the covertness of the attack, their performance is generally limited due to the lack of specialized adjustment catered to the target model. In this work, we propose the self-ensemble strategies to address the issue and greatly boost the black-box attack transferability of the existing transfer-based detector attack.

\subsection{Object Detectors}

Object detection is a fundamental computer vision technology predicting objects' categories and positions and is widely used in many perception tasks. In the past ten years, deep learning-based models have greatly improved the performance of object detectors. The mainstream methods based on deep learning can be roughly divided into one-staged and two-staged methods. The former directly predicts the positions and classes of the object instances and are thus faster; the latter first use the region proposal network (RPN) to generate proposals, and then predict the labels and positions of the selected proposals. In this paper, we select eight mainstream detectors of both one-stage and two-stage, including YOLO v2~\cite{redmon2017yolo9000}, YOLO v3 \& YOLO v3tiny~\cite{redmon2018yolov3}, YOLO v4 \& YOLO v4tiny ~\cite{bochkovskiy2020yolov4}, YOLO v5~\cite{glenn_jocher_2020_4154370}, Faster R-CNN~\cite{ren2015faster} and SSD~\cite{liu2016ssd} to systemically verify the proposed T-SEA framework.

\subsection{Attacks on Object Detector}

The security of deep learning-based models is receiving increasing attention~\cite{chen2022towards, chen2022shape}, especially with the existence of adversarial attacks. Recent works have explored adversarial robustness of object detectors. To begin, \cite{xie2017adversarial} apply adversarial attack on object detectors, performing iterative gradient-based method to misclassify the proposals, and the similar ideas are also carried out by \cite{li2018robust, wei2019transferable}. Ensuingly, for enhancing the attack capability in the physical world, \cite{thys2019fooling, chen2018shapeshifter} propose real-world adversarial patches to attack the mainstream detectors, \eg, YOLO and Faster R-CNN. Recently, model ensemble approaches are used to improve the attack transferability among multiple detectors, such as \cite{huang2021rpattack, zhao2020object, wu2020dpattack} simultaneously attacking multiple detectors to generate cross-model adversarial examples. Different from the above, we focus on how to make the most of the limited information (i.e., a single detector) to carry out a high-transferability black-box attack on multiple black-box detectors.

\section{Method}
\label{sec:method}

In this section, we first give the problem formulation in Sec.~\ref{method:problem_formulation} and describe the overall framework in Sec.~\ref{method:overall_framework}. Then, we respectively introduce the details of our enhanced baseline and self-ensemble strategies in Sec.~\ref{method:enhanced_baseline} and Sec.~\ref{method:self_ensemble_strategies}.

\subsection{Problem Formulation}
\label{method:problem_formulation}


In this work, we carry out a transfer-based black-box attack with only one white-box detector to decrease the mean average precision (mAP) of both white-box and black-box detectors. Given the target input data distribution $\mathcal{D}(\mathcal{X}, \mathcal{F})$, we regard a single pre-trained detector $f_w \in \mathcal{F}$ as the white-box attacked model, $x_{1,..,N}\in \mathcal{X}$ as the input images, where $N$ is the number of training samples. We are committed to crafting a universal adversarial patch $\tau$ from the adversarial distribution $\mathcal{T}$ to disrupt the detection process, 
\begin{equation}
    \begin{aligned}
    \hat{\tau} = \mathop{\arg\min}_{\tau \sim \mathcal{T}} \sum_{i}^N L (\widetilde{x_i}, \widetilde{f}_w, \widetilde{\tau}) + \lambda J(\tau), \\
    \end{aligned}
\end{equation}
where $L$ is the loss function to measure the corruption of detector, $\widetilde{x_i}, \widetilde{f}_w$ and $\widetilde{\tau}$ denote self-ensembled data, model and patch, respectively, and $J(\cdot)$ is a regularization term which we employ total variation of $\tau$ in our work.

\subsection{Overall Framework}
\label{method:overall_framework}

As depicted in Fig.~\ref{fig:pipeline}, we divide the entire T-SEA pipeline into training stage and attack stage. During the training stage, the input images are first augmented with the constrained data augmentation. Ensuingly, the white-box detector (i.e., the attacked detector) is used to locate the target objects from the augmented images. Then, we attach adversarial patches to the center of each detected object of the target class, along with patch cutout for mitigating overfitting. After that, the images with adversarial patches go through the shake-dropped models, and we minimize the object confidence and continuously optimize the training adversarial patches until reaching maximum epoch number. During the attack stage, we apply our well-optimized adversarial patch on the test images to disrupt the detection process of multiple black-box detectors.

\subsection{Enhanced Baseline}
\label{method:enhanced_baseline}

In this section, we introduce the details of the enhanced baseline, which is described in Algorithm~\ref{alg1}. Firstly, we randomly initialize a patch $\tau_0$, prepare the training images and attacked model $f_w$. During each training epoch, for every input image batch $X$, we first obtain their detection results via the white-box detector $f_w$. Ensuingly, we utilize the transformation function $T$ to apply the training patch into each image to generate adversarial image batch $X^{adv}$. Next, we calculate the object confidence of $X^{adv}$, and the attack loss of the image batch. Here we use $\ell_1$ loss as the attack loss, unless otherwise stated. Finally, we update the $\tau$ via base attack method $\hbar$ and adjust the learning rate after each epoch until reaching the maximum number.

\begin{algorithm}[htb]
\caption{Enhanced Baseline Based on \cite{thys2019fooling}} 
\label{alg1}
\begin{algorithmic}[1]

\REQUIRE ~~
 $x_{1,..,N}$ (training images), $f_w$ (white-box detector), $M$ (maximum epoch), $BS$ (batch size), $\tau_0$ (the initial patch), $T$ (patch applier), $\hbar$ (base attack method), $scheduler$ (learning rate scheduler).

\ENSURE ~~
Well-optimized Adversarial Patch $\tau$
\STATE $\tau \Leftarrow \tau_0$

\FOR{each $i \in [1, M]$}
\FOR{each $j \in [1, \frac{N}{BS}]$}
\STATE $X \Leftarrow x_{(j-1)\cdot BS+1}, \ ...,\ x_{j\cdot BS}$
\STATE $bbox^{clean}, \ conf^{clean} \Leftarrow f_w(X)$
\STATE $X^{adv} \Leftarrow T(X, \ bbox^{clean}, \ \tau)$
\STATE $bbox^{adv}, \ conf^{adv} \Leftarrow f_w(X^{adv})$
\STATE $loss \Leftarrow Avg(conf^{adv})$
\STATE $\tau \Leftarrow \hbar(\tau, \ loss)$
\ENDFOR
\STATE update $lr$ via $scheduler$
\ENDFOR


\end{algorithmic}
\end{algorithm}

Compared to AdvPatch~\cite{thys2019fooling}, we slightly adjust two training strategies. One is the learning rate scheduler. We observe that a saddle point during patch optimization may lead the original plateau-based scheduler to drastically drop the learning rate, causing inadequate optimization. Hence, we adjust the learning rate decline strategy with the following criterion,
\begin{eqnarray}
	lr \Leftarrow lr * \mu, if \ (\ell_t-\ell_{t-1}) < \epsilon_1 \ and \ \frac{(\ell_t-\ell_{t-1})}{\ell_t} < \epsilon_2,
\end{eqnarray}
where $\mu$ is the decay factor, $\ell_t$ denotes the mean loss at epoch $t$, $\epsilon_1, \epsilon_2$ are thresholds to control learning rate updates. In our experiments, we ensure that the learning rate decrease more stably via the above hyper-parameters. The other adjustment is to reduce the patch scale at training stage $s_t$. Here, the patch scale is the length ratio between the patch and the bounding box of the specific object. As Fig.~\ref{fig:scheduler} shows, a proper scale helps the optimized patch to learn more global patterns, which will not be disrupted easily when the patch is scaled, \eg, a local pattern will lose more information when its corresponding patch is scaled to a very small size.

\subsection{Self-Ensemble Strategies}
\label{method:self_ensemble_strategies}

\subsubsection{Theoretical Analysis}

Typical model training usually attempts to find a mapping function $f$ from a finite hypothesis space $\mathcal{F}$ that describes the relationship between the input data $x$ and label $y$ following the underlying joint distribution $\mathcal{D}(\mathcal{X}, \mathcal{Y})$. 
Generally, since $\mathcal{D}$ is unknown, we use training data set $S=\{(x_i, y_i)|i= 1,...,N\}$ to train the model, which means the optimization involves the empirical risk minimization(ERM)~\cite{vapnik1999nature}:
\begin{equation}
    \widehat{R}_S(f) = \frac{1}{N}\sum_{i}^N L(y_i, f(x_i)),
\end{equation}
where $L$ is a loss function to measure the difference between the model prediction $f(x)$ and the label $y$. However, the empirical risk is unable to provide generalization on unseen data~\cite{zhang2017mixup}. Fortunately, we can derive that the generalization error $R_{\mathcal{D}}(f)$ is bounded by $\widehat{R}_S(f)$ for a given confidence $1-\sigma \in (0, 1)$~\cite{mohri2018foundations, belkin2018understand},
\begin{equation}
    \forall f \in \mathcal{F}, \ R_{\mathcal{D}}(f) \leq \widehat{R}_S(f) + \sqrt{\frac{log \ c + log \ \frac{1}{\sigma}}{N}},
\end{equation}
where $N$ denotes the input data size, $c$ is a complexity measure of $\mathcal{F}$, such as VC-dimension~\cite{mohri2018foundations} or covering numbers~\cite{anthony1999neural}. This generally yields bounds of the generalization gap $R_\mathcal{D}(f)-\widehat{R}_S(f)=\mathcal{O}(\sqrt{ \frac{log \ c}{N}})$, which gives us a lens to analyze model generalization: we can sample more training data or lower model complexity to help close the generalization gap. In practice, limited training data and large-scaled model parameters make the model prone to overfit~\cite{ying2019overview}, causing an undesirable generalization gap increase. Fortunately, past studies have widely adopted a series of regularization strategies on both input data and the model, such as data augmentation~\cite{zhang2017mixup, devries2017improved, zhong2020random}, Stochastic Depth~\cite{stochastic_depth}, and Dropout~\cite{srivastava2014dropout}.

Motivated by the aforementioned model regularization methods, we discuss how to improve attack transferability of the adversarial patch. During patch optimization, the model $f$ becomes an input instead of the optimization objective, which means the original input distribution $\mathcal{D}(\mathcal{X}, \mathcal{Y})$ is shifted to a joint input distribution $\mathcal{D}(\mathcal{X}, \mathcal{F})$. Naturally, we can improve generalization of the adversarial patch by increasing training data scale $N_{x,f}$, i.e. sampling more training images $x$ and ensembling more white-box models $f_w$. However, obtaining massive images or models of the same task is generally expensive or impractical for the attacker. To address this issue, we propose to virtually expand the input by data augmentation and model ShakeDrop to help improve patch transferability. Meanwhile, although the large-scale models have achieved superior performance in most tasks, these models still benefit from reducing model complexity at training stage like stochastic depth~\cite{stochastic_depth} and Dropout~\cite{srivastava2014dropout}. Thus we propose the patch cutout to reduce its capacity $c_\tau$ in training process as a new regularization to alleviate overfitting.

\subsubsection{Data: Constrained Data Augmentation}

Motivated by data augmentation in increasing model generalization, we can similarly improve patch transferability by virtually expanding the training set $S$ to more closely approximate the underlying distribution $\mathcal{D}$. Therefore, we employ a constrained policy to avoid unnatural augmentation which may not appear in natural scenes. In our work, we 1) mildly resize and crop the input images, 2) slightly alter the brightness, contrast, saturation and hue, and 3) randomly rotate the input images in a small range, to generate natural augmented images. 

\subsubsection{Model: ShakeDrop}

\begin{figure}[h]
    \centering
    \subfloat[Forward\label{shakedrop-forward}]{%
        \includegraphics[width=0.4\linewidth]{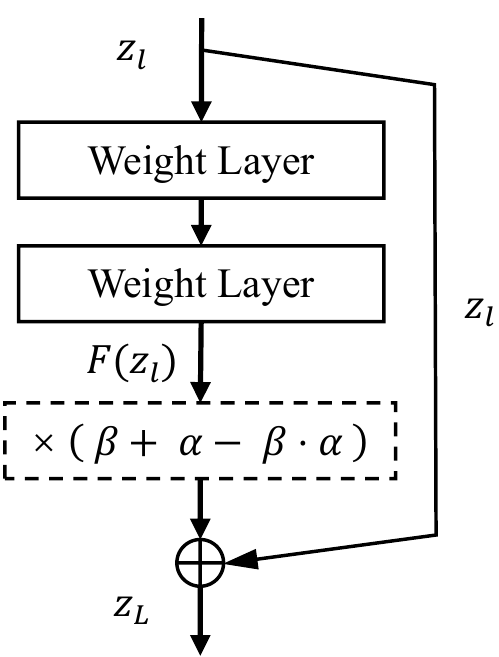}}
    \quad
    \quad
    \subfloat[Backward\label{shakedrop-backward}]{%
        \includegraphics[width=0.4\linewidth]{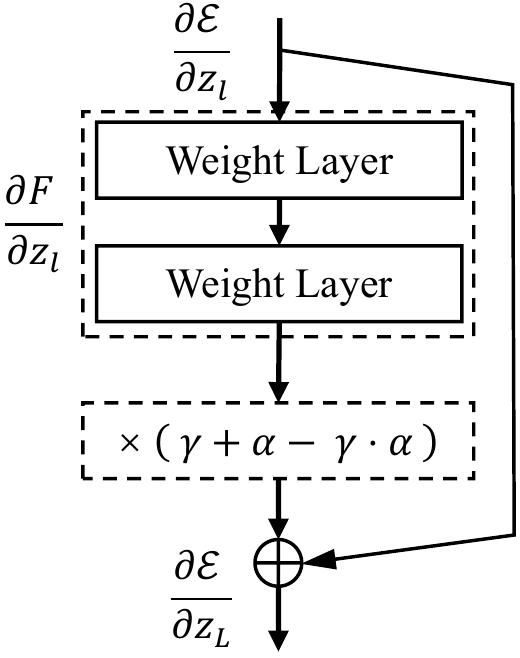}}
    \caption{We utilize the model ShakeDrop to linearly combine the outputs of the stacked layers and identity of a residual block.}
  \label{fig:model-shakedrop}
\end{figure}

As discussed above, when training the patch, sampling more white-box models $f$ from the joint input distribution $\mathcal{D}$ can improve the adversarial transferability of the crafted patch. However, it is difficult to obtain multiple models of the same task in reality. Inspired by stochastic depth~\cite{stochastic_depth}, which randomly drops a subset of layers to virtually ensemble multiple model variants for improving generalization, we utilize ShakeDrop~\cite{yamada2018shakedrop} to linearly combine the outputs of stacked layers and identity of a residual block, generating massive variants of the attacked white-box model as the Fig.~\ref{fig:model-shakedrop} illustrated.

Specifically, in forward propagation, we combine the identity $z_l$ and the output of stacked layers $F(z_l)$ of a residual block to generate $z_L$ with the following formula:
\begin{equation}
\begin{split}
    z_L =  z_l + (\beta + \alpha - \beta \cdot \alpha) \cdot F(z_l),
\end{split}
\end{equation}
where $\beta$ is sampled from a Bernoulli distribution with $P(\beta=1)=\varphi_s$, $\alpha$ is sampled from a continuous uniform distribution $\alpha \sim U(1-e, 1+e)$ ($e$ is a constant). In backward propagation, denoting the loss function as $\mathcal{E}$, from the chain rule of backpropagation, ShakeDrop can be formulated as
\begin{align}
    \begin{aligned}
    \psi(\frac{\partial\mathcal{E}}{\partial z_l}) &=
    \psi(\frac{\partial\mathcal{E}}{\partial z_L} \cdot \frac{\partial z_L}{\partial z_l}) \\ 
    &=\frac{\partial\mathcal{E}}{\partial z_L} (1+(\gamma+\alpha - \gamma \cdot \alpha) \cdot \frac{\partial F}{\partial z_l}),
    \end{aligned}
\end{align}
where $\gamma$ is another Bernoulli variable and identically distributed with $\beta$.

\subsubsection{Patch: Cutout}

Dropout~\cite{srivastava2014dropout} is designed to randomly drop neural units along with their connections, effectively preventing overfitting of the model. Inspired by Dropout, we propose a patch cutout strategy, randomly masking a region of the adversarial patch to reduce the patch complexity at training stage and thus to prevent overfitting on specific model and data. From an other angle, model Dropout prevents excessive co-adapting of neural units and spread out features over multi neurons to alleviate overfitting. Similarly, patch cutout also prevents the adversarial nature of patch from relying too much on the patterns of a certain area. 

The proposed patch cutout is similar to cutout~\cite{devries2017improved} or random erasing~\cite{zhong2020random} in model training. Specifically, for a normalized input image $I_{x, y} \in [0, 1]$ of size $H \times W$, we carry out the following process with a probability $\varphi_c$ before attaching patches to the target objects: 1) we firstly conduct random sampling to obtain one point $p=(x_0, y_0)$ within the given patch; 2) then we cover a $\eta H \times \eta W$ square area($\eta$ is the ratio) centered at $p$ with a specific value $k \in [0, 1]$.

\section{Experiment}
\label{sec:experiment}

\begin{table*}[]
\setlength\tabcolsep{3pt}
\centering
\small
\begin{tabular}{@{}c|cccccccc|c@{}}
\toprule
\textbf{Methods} & YOLO v2 & YOLO v3 & YOLO v3tiny & YOLO v4 & YOLO v4tiny & YOLO v5 & Faster R-CNN & SSD & \textbf{Black-Box Avg$\downarrow$}  \\ \midrule
AdvPatch   & \textbf{5.66} & 40.26 & 18.07 & 48.49 & 24.44 & 43.38 & 39.27 & 41.28	& 36.46   \\
T-SEA(ours)    & \textbf{1.73}	& 4.48	& 2.41	& 5.68	& 7.75	& 6.91	& 16.38	& 20.55	& 9.16 $^{27.30\downarrow}$ \\ \midrule
AdvPatch  & 51.85 & \textbf{13.89} & 51.17 & 57.16 & 58.43 & 70.47 & 51.46 & 59.79 & 57.19  \\
T-SEA(ours)                                                                 & 31.02 & \textbf{5.76}  & 35.38 & 42.37 & 25.3  & 58.02 & 37.62 & 52.26 & 40.28 $^{16.91\downarrow}$  \\ \midrule
AdvPatch     & 56.02 & 66.12 & \textbf{1.39} & 69.64 & 51.56 & 72.16 & 56.61 & 60.73 & 61.83   \\
T-SEA(ours)    & 38.85 & 47.13 & \textbf{0.51} & 52.75 & 31.01 & 61.18 & 49.12 & 55.44 & 47.93 $^{13.90\downarrow}$  \\ \midrule
AdvPatch & 37.41 & 37.18 & 17.58 & \textbf{19.67} & 26.91 & 46.37 & 44.67 & 43.07 & 36.17  \\
T-SEA(ours)                                & 9.34  & 6.46  & 9.49  & \textbf{4.22}  & 16.65 & 11.66 & 16.24 & 29.43 & 14.18 $^{21.99 \downarrow}$   \\ \midrule
AdvPatch     & 47.41 & 59.59 & 37.14 & 66.48 & \textbf{14.50} & 69.51 & 55.95 & 55.22 & 55.90   \\
T-SEA(ours)    & 37.69 & 44.98 & 15.81 & 51.14 & \textbf{4.11}  & 51.37 & 46.06 & 49.39 & 42.35 $^{13.55\downarrow}$  \\ \midrule
AdvPatch  &  46.9  & 54.93 & 29.20  & 62.16 & 46.15 & \textbf{13.39} & 47.71 & 50.73 & 48.25   \\
T-SEA(ours)    & 11.61 & 16.77 & 10.73 & 32.53 & 17.55 & \textbf{1.37}  & 17.14 & 30.78 & 19.59 $^{28.66\downarrow}$  \\ \midrule
AdvPatch     & 24.53 & 23.37 & 14.58 & 26.54 & 25.6  & 30.46 & \textbf{8.62} & 42.15 & 26.75   \\
T-SEA(ours)    & 9.28  & 4.08  & 3.99          & 8.55  & 13.58 & 10.45 & \textbf{3.08} & 26.46 & 10.91 $^{15.84\downarrow}$  \\ \midrule
AdvPatch     & 32.43 & 62.93 & 52.03 & 66.22 & 49.07 & 51.19 & 47.42 & \textbf{15.10}  & 51.61   \\
T-SEA(ours)    & 12.06 & 30.90 & 9.73          & 27.33 & 9.08  & 17.38 & 28.54 & \textbf{5.13}  & 19.29 $^{32.32\downarrow}$  \\ \midrule

\end{tabular}
\caption{Comparisons between T-SEA and AdvPatch~\cite{thys2019fooling}. We attack each detector separately and use the remaining seven detectors as the black-box detectors. The proposed T-SEA performs much better than \cite{thys2019fooling} on both white-box setting and black-box setting, demonstrating the effectiveness of the proposed self-ensemble strategies.}
\label{exp:results_on_different_attack_models}
\vspace{-5mm}
\end{table*}

In this section, we first introduce the implementation details in Sec.~\ref{epx:Datasets_and_Implementation_Details}. Then, we present the main results of T-SEA in Sec.~\ref{exp:Main_Results}, reporting its performance on different detectors and attack methods. After that, we compare the proposed method with other state-of-the-art detection attack algorithms in Sec.~\ref{exp:Comparison_with_SOTA_Methods} and perform the ablation study in Sec.~\ref{exp:Ablation_Study}. Finally, we show the transferability of T-SEA across the datasets and scenes in Sec.~\ref{exp:across_datasets_verification} and Sec.~\ref{exp:physical_verification}.

\begin{table}[]
\setlength\tabcolsep{2.5pt}
\centering
\small
\begin{tabular}{@{}cc|c|c@{}}
\toprule
\multicolumn{2}{c|}{Method} & White Box $\downarrow$  & Black-Box Avg$\downarrow$ \\ \midrule

\multirow{2}{*}{Adam} & AdvPatch   & 13.39 & 48.25  \\
                     & T-SEA(ours) & 1.37  & 19.59  \\ \midrule
\multirow{2}{*}{SGD} & AdvPatch    & 20.39 & 55.97  \\
                     & T-SEA(ours) & 1.66  & 15.85  \\ \midrule
\multirow{2}{*}{MIM}  & AdvPatch   & 11.91 & 33.51  \\
                     & T-SEA(ours) & 1.43  & 15.43  \\ \midrule
\multirow{2}{*}{BIM}  & AdvPatch   & 8.47  & 32.93  \\
                     & T-SEA(ours) & 1.62  & 19.15  \\ \midrule
\multirow{2}{*}{PGD}  & AdvPatch   & 13.58 & 38.70  \\
                     & T-SEA(ours) & 1.60  & 18.06  \\ \midrule
\end{tabular}
\caption{Comparisons of T-SEA and AdvPatch with Different Base Attack Methods on YOLO v5 (white-box). We select five classical base attack methods, including optimization-based methods (Adam and SGD) and iterative methods (MIM, BIM and PGD). The results show that T-SEA can enhance the performance of all these methods and performs much better than AdvPatch.}
\label{exp:results_of_different_attack_methods}
\vspace{-4mm}
\end{table}

\subsection{Implementation Details}
\label{epx:Datasets_and_Implementation_Details}

\textbf{Datasets} In our experiments, we utilize the INRIA person dataset~\cite{dalal2005histograms} to train and test our adversarial patch, whose training set and test set consist of 614 and 288 images, respectively. Meanwhile, to verify the transferability of the crafted patch, we select images containing person from COCO validation set (named COCO-person) and CCTV Footage of Humans\footnote{https://www.kaggle.com/datasets/constantinwerner/human-detection-dataset} (named CCTV-person) as additional test data. The former contains 1684 human images in different scenes (e.g., sports playground, transportation routes, oceans and forests) and the latter contains 559 in-person images from camera footage.

\textbf{Optimization Details} We use INRIA train set as the training set and regard person as the target attack class. The patch size is $300\times300$, the input image size is $416\times416$, the batch size $BS=8$, and the maximum epoch number $M=1000$. For the E-baseline, we adjust the training patch scale from $0.2$ in AdvPatch~\cite{thys2019fooling} to $0.15$, set $\epsilon_1=1e-4, \epsilon_2=1e-4$ for the learning rate scheduler, and adopt the Adam~\cite{kingma2014adam} as the optimizer. For constrained data augmentation, we carry out constrained data augmentation via horizontal flip, slight color jitter, random resized crop, and random rotation; for model ShakeDrop, the constant $e$ is set to $1$ so that $\alpha \sim U(0,2)$, and we perturb the model with the probability of $\varphi_s=0.5$; for patch cutout, we set the fill value of the erased area $k=0.5$, the ratio $\eta=0.4$, and the probability $\varphi_c=0.9$.

\textbf{Evaluation Metric} Following~\cite{hu2021naturalistic, thys2019fooling}, we use the mean Average Precision (mAP) to measure the attack capability of the crafted patch (the lower mAP, the better attack) and regard the detector's predictions of the clean data as the ground truth (i.e., $mAP=1$).

\begin{table*}[]
\setlength\tabcolsep{1.5pt}
\centering
\small
\begin{tabular}{@{}c|c|ccccccc|c@{}}
\toprule
\multirow{2}{*}{{\textbf{Method}}} & White Box $\downarrow$ & \multicolumn{7}{c|}{Black Box $\downarrow$} & \multirow{2}{*}{\textbf{Black-Box Avg$\downarrow$}} \\ \cmidrule(r){2-9}
 & YOLO v2 & YOLO v3 & YOLO v3tiny & YOLO v4 & YOLO v4tiny & YOLO v5 & Faster R-CNN & SSD &  \\ \midrule
Gray   & 67.75 & 76.22 & 80.69 & 75.22 & 76.89 & 81.86 & 61.75 & 72.05 & -  \\
Random Noise & 70.67 & 75.8  & 82.44 & 75.1  & 78.74 & 81.79 & 63.41 & 72.9  & - \\
White  & 68.52 & 74.89 & 80.2  & 74.73 & 76.09 & 80.09 & 60.35 & 69.41  & - \\ \bottomrule

NPAP~\cite{hu2021naturalistic}     & \textbf{38.03} & 56.85 & 58.04         & 67.74 & 67.43 & 66.85 & 56.6  & 56.66 & \textbf{61.45}   \\
AdvCloak~\cite{wu2020making}     & \textbf{33.74} & 54.77 & 53.42         & 67.57 & 56.12 & 68.05 & 55.19  & 60.82 & \textbf{59.42}   \\

AdvPatch~\cite{thys2019fooling}  & \textbf{5.66} & 40.26 & 18.07 & 48.49 & 24.44 & 43.38 & 39.27 & 41.28 & \textbf{36.46}  \\ 
E-baseline(ours)   & \textbf{3.61} & 15.32 & 5.50 & 18.58 & 13.39 & 9.77 & 29.73 & 23.82 & \textbf{16.59} \\
T-SEA(ours)  & \textbf{1.73}	& 4.48	& 2.41	& 5.68	& 7.75	& 6.91	& 16.38	& 20.55	& \textbf{9.16}  \\ \bottomrule
\end{tabular}
\caption{Comparisons with Existing Detection Attack Methods. For the clearer controlled observations, we list the results of gray, random noise, and white patch. Compared with existing methods, the proposed T-SEA achieves the best performance on both white-box attack and black-box attack.}
\label{exp:comparison_with_SOTA_methods}
\end{table*}

\subsection{Main Results}
\label{exp:Main_Results}

\subsubsection{Results on Different Attack Models}

We systematically investigate attack performance of the proposed T-SEA on eight widely-used, object detectors, and report the quantitative results in Tab.~\ref{exp:results_on_different_attack_models}. Since T-SEA is improved based on AdvPatch~\cite{thys2019fooling}, we also report the results of AdvPatch. The proposed T-SEA achieves significant improvements on both white-box and black-box performance of all eight detectors compared to \cite{thys2019fooling}, demonstrating the effectiveness of the proposed strategies. Meanwhile, for some white-box detectors (\eg, YOLO v2 and Faster R-CNN), the black-box average mAP can drop to around 10, exhibiting effective black-box performance of T-SEA.

\subsubsection{Results of Different Attack Methods}

T-SEA is not designed only for a specific base attack method; it is important for T-SEA to perform well on different base attack methods. In Tab.~\ref{exp:results_of_different_attack_methods}, we give the detailed results of T-SEA with different base attack methods on YOLO v5, which includes the optimization-based methods (e.g., Adam and SGD) and iterative methods (e.g., MIM, BIM and
PGD). The results show that the proposed T-SEA can improve both white-box and black-box attack performance on all methods above, demonstrating that the performance gain caused by T-SEA is not limited to a specific method. That is to say, T-SEA may potentially work well with future base attack methods to further increase the transferability of their crafted AdvPatch.

\subsection{Comparison with SOTA Methods}
\label{exp:Comparison_with_SOTA_Methods}

In this section, we compare T-SEA with the SOTA detection attack approaches, which all regard the YOLO v2 as the white-box model and evaluate on seven black-box detectors, and we also report the results of gray/random noise/white patches as control group. As reported on Tab.~\ref{exp:comparison_with_SOTA_methods}, 1) all adversarial patch performs much better than the control group; 2) our E-baseline and T-SEA achieve the highest performance among all adversarial approaches 
with the same inference setting (i.e., we ensure the perturbed area is same for each method), showing that compared to the existing single model attack method, the training strategies adjustment and self-ensemble strategies can effectively enhance the attack capability on both white-box and black-box of the crafted adversarial patch.

\subsection{Ablation Study}
\label{exp:Ablation_Study}

\subsubsection{Training Strategies Adjustment}

\begin{table}[!t]
    \centering
    \setlength\tabcolsep{2pt}
    \begin{tabular}{lcc}
    \toprule
     & \textbf{White Box} $\downarrow$ &  \textbf{Black Box} $\downarrow$ \\ \midrule
    E-baseline & 13.39 & 48.25   \\
    + Constrained Data Aug. & 18.47 & 42.42  \\
    + Model ShakeDrop & 12.54 & 39.40  \\
    + Patch cutout & 6.80 & 32.54  \\
    + Combined Strategies & 1.37 & 19.59  \\
    \bottomrule
    \end{tabular}
    \vspace{-1mm}
    \caption{Ablation Study of Self-Ensemble Strategies on YOLO v5 (white-box). Each self-ensemble strategy is added to the E-baseline to verify its individual performance gain. The results show that: 1) though using data augmentation alone degrades the white-box performance, all strategies can obviously improve the black-box performance of E-baseline; 2) combining all achieves non-trivial results on both white-box and black-box setting.}
    \label{ablation:self_ensemble_strategies}
    \vspace{-1mm}
\end{table}

\begin{figure}[t]
  \centering
  \includegraphics[width=.95\linewidth]{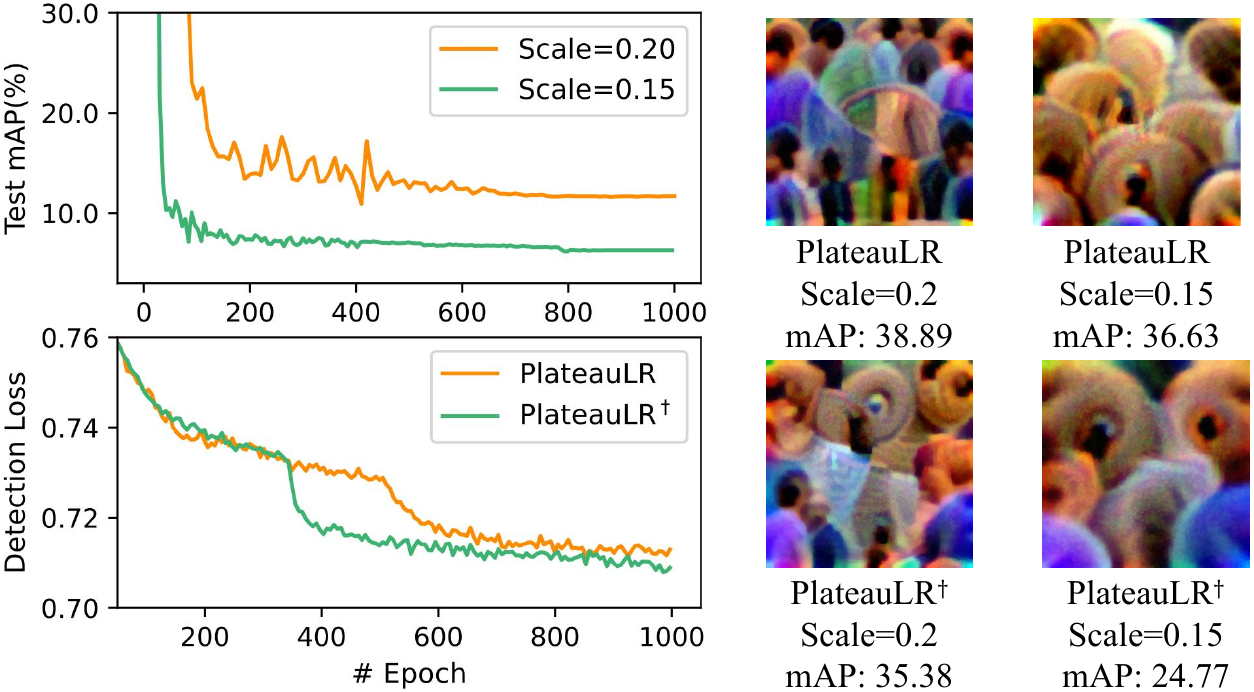}
  \caption{The improvements after adjusting learning rate scheduler (tagged as PlateauLR$^\dagger$) and patch scale at training stage. A large patch scale during training will cause the test mAP to drop lower and faster, and the adjusted PlateauLR$\dagger$ can also cause the detection loss to further decrease.}
  \label{fig:scheduler}
\end{figure}

Here we explore the performance gain from the training strategies adjustments. As discussed in the Sec.~\ref{method:enhanced_baseline}, we modify the learning rate scheduler and training patch scale of \cite{thys2019fooling}. As illustrated in Fig.~\ref{fig:scheduler}, the modified training scale can reduce the mAP in test set more quickly and effectively, and the crafted patch has clearer adversarial patterns. The adjusted scheduler can also lead the detection loss to decrease more, and can be further combined with the modified training scale to achieve a better result.

\subsubsection{Self-Ensemble Strategies}

In this work, we are motivated by model training approaches that enhance generalization ability, and we propose self-ensemble strategies on the input data, the attacked model, and the training patch, greatly augmenting adversarial transferability from themselves. Here we individually inspect the proposed strategies on our E-baseline (YOLO v5 as white-box detector) to explore the impact of each strategy. As reported in Tab.~\ref{ablation:self_ensemble_strategies}, though the constrained data augmentation will slightly decrease the white-box performance (other two strategies both improve the white-box performance), all these self-ensemble strategies increase the black-box results separately. and combining them can achieve the best performance.

\subsection{Cross-Dataset Verification}
\label{exp:across_datasets_verification}

\begin{table}[]
\setlength\tabcolsep{3pt}

\begin{tabular}{cccc}
\toprule
                        \textbf{Dataset}   & \textbf{Method}  & \textbf{White Box} $\downarrow$ & \textbf{Black Box} $\downarrow$ \\ \midrule
\multirow{2}{*}{COCO-person}     & AdvPatch~\cite{thys2019fooling}  & 45.83       & 52.54      \\
                                 & T-SEA     & 37.28       & 38.87      \\ \midrule
\multirow{2}{*}{CCTV-person}     & AdvPatch~\cite{thys2019fooling}  & 38.07      &  34.08      \\ 
                                 & T-SEA     & 38.71      & 19.91       \\ \bottomrule
\end{tabular}
\caption{The Cross Datasets Transferability of T-SEA on YOLO v5 (white-box). Compared to the \cite{thys2019fooling}, the adversarial patch crafted by T-SEA has much stronger attack adaptability on person images from different datasets.}
\label{ablation:across_datasets_verification}
\end{table}

The capability of performing cross-dataset attack is also significant for a well-optimized patch. Here, we apply the INRIA-trained patch on two different datasets, COCO-person and CCTV-person. Compared to INRIA, COCO-person has much more person images on different scenes, while CCTV-person focus on persons from the security camera. As shown in Tab.~\ref{ablation:across_datasets_verification}, the patch crafted by T-SEA achieves comparable results on the white-box setting with the AdvPatch, but obtains a much better black-box attack capability on both COCO-person and CCTV-person, indicating its strong black-box cross-data attacking capability.

\subsection{Physical Verification}
\label{exp:physical_verification}

\begin{figure}[t]
  \centering
  \includegraphics[width=.95\linewidth]{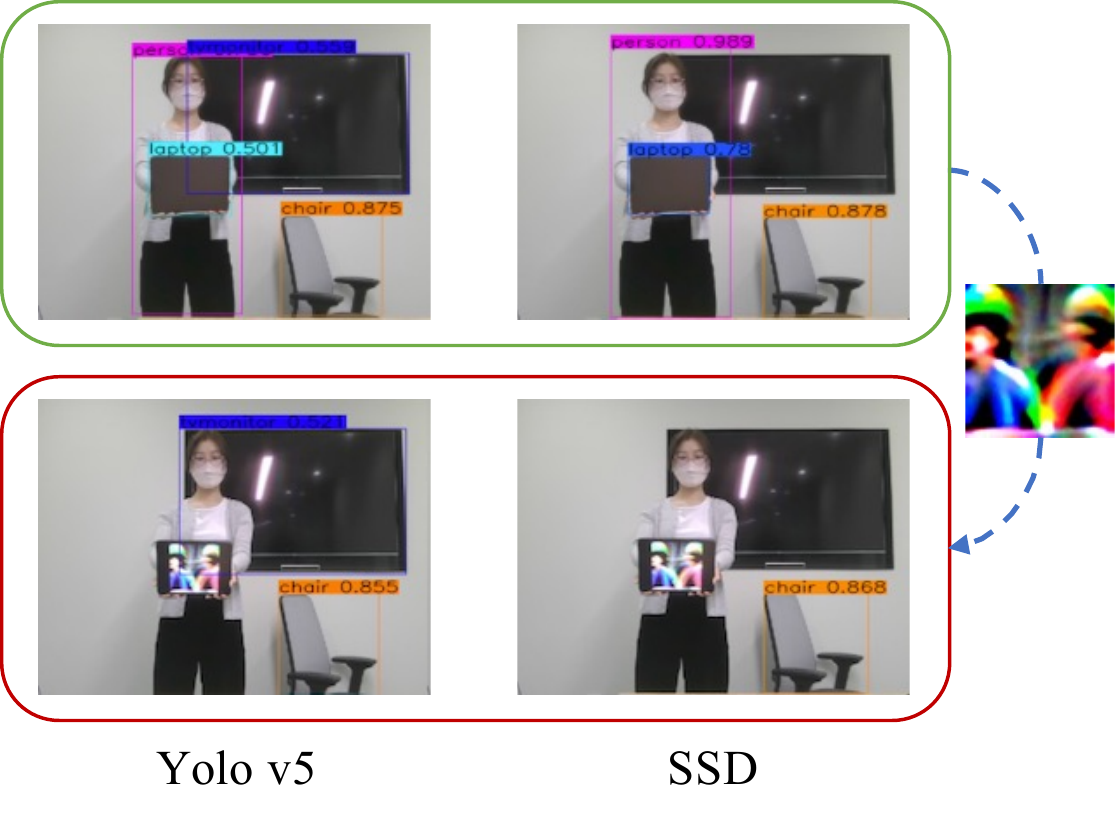}
  \caption{Physical Attack Demo of T-SEA. After showing the optimized patch on iPad, the YOLO v5 and SSD can not detect person.}
  \label{fig:physical}
\end{figure}

Although applying the patch attack in the physical world is not the main goal of our work, we believe the optimized patch that has high transferability may perform well in physical settings. Here we show a simple case in Fig.~\ref{fig:physical} that the patch shown on a iPad can successfully attack the human detection of YOLO v5 and SSD, without disrupting the detection process of other objects.

\section{Future Work}

Since most existing mainstream detectors are CNN-based, we focus on applying the self-ensemble strategies on these detectors in this work. However, the transformer-based detectors~\cite{carion2020end, liu2021swin} are achieving promising results, and the proposed model ShakeDrop can not directly applied on transformers (the others can), we will follow the motivation of designing the model ShakeDrop, to propose new approach and generate variants of transformer-based detector.
\section{Conclusion}
\label{sec:conclusion}

In this paper, we propose a novel transfer-based self-ensemble black-box attack on object detectors, achieving stable and excellent performance gains with various base attack methods on multiple popular object detectors. Firstly, with only slight training strategy adjustments, we improve the existing method's performance and regard it as our enhanced baseline. Then, based on this baseline, we propose a series of self-ensemble strategies to augment the input data, the attacked model, and the training patch from itself to significantly enhance the adversarial transferability of the optimized patch on black-box detectors. The comprehensive experimental results reveal the potential risk that an attacker with only one model could still succeed in a high transferability black-box attack.


{\small
\bibliographystyle{ieee_fullname}
\bibliography{egbib}

\begin{thebibliography}{10}\itemsep=-1pt

\bibitem{anthony1999neural}
Martin Anthony and Peter Bartlett.
\newblock {\em Neural network learning: Theoretical foundations}.
\newblock cambridge university press, 1999.

\bibitem{bai2020adversarial}
Song Bai, Yingwei Li, Yuyin Zhou, Qizhu Li, and Philip~HS Torr.
\newblock Adversarial metric attack and defense for person re-identification.
\newblock {\em IEEE Transactions on Pattern Analysis and Machine Intelligence},
  43(6):2119--2126, 2020.

\bibitem{belkin2018understand}
Mikhail Belkin, Siyuan Ma, and Soumik Mandal.
\newblock To understand deep learning we need to understand kernel learning.
\newblock In {\em International Conference on Machine Learning}, pages
  541--549. PMLR, 2018.

\bibitem{bochkovskiy2020yolov4}
Alexey Bochkovskiy, Chien-Yao Wang, and Hong-Yuan~Mark Liao.
\newblock Yolov4: Optimal speed and accuracy of object detection.
\newblock {\em arXiv preprint arXiv:2004.10934}, 2020.

\bibitem{brendel2017decision}
Wieland Brendel, Jonas Rauber, and Matthias Bethge.
\newblock Decision-based adversarial attacks: Reliable attacks against
  black-box machine learning models.
\newblock {\em arXiv preprint arXiv:1712.04248}, 2017.

\bibitem{carion2020end}
Nicolas Carion, Francisco Massa, Gabriel Synnaeve, Nicolas Usunier, Alexander
  Kirillov, and Sergey Zagoruyko.
\newblock End-to-end object detection with transformers.
\newblock In {\em European conference on computer vision}, pages 213--229.
  Springer, 2020.

\bibitem{chen2018shapeshifter}
Shang-Tse Chen, Cory Cornelius, Jason Martin, and Duen Horng~Polo Chau.
\newblock Shapeshifter: Robust physical adversarial attack on faster r-cnn
  object detector.
\newblock In {\em Joint European Conference on Machine Learning and Knowledge
  Discovery in Databases}, pages 52--68. Springer, 2018.

\bibitem{chen2022shape}
Zhaoyu Chen, Bo Li, Shuang Wu, Jianghe Xu, Shouhong Ding, and Wenqiang Zhang.
\newblock Shape matters: deformable patch attack.
\newblock In {\em European Conference on Computer Vision}, pages 529--548.
  Springer, 2022.

\bibitem{chen2022towards}
Zhaoyu Chen, Bo Li, Jianghe Xu, Shuang Wu, Shouhong Ding, and Wenqiang Zhang.
\newblock Towards practical certifiable patch defense with vision transformer.
\newblock In {\em Proceedings of the IEEE/CVF Conference on Computer Vision and
  Pattern Recognition}, pages 15148--15158, 2022.

\bibitem{dalal2005histograms}
Navneet Dalal and Bill Triggs.
\newblock Histograms of oriented gradients for human detection.
\newblock In {\em 2005 IEEE computer society conference on computer vision and
  pattern recognition (CVPR'05)}, volume~1, pages 886--893. Ieee, 2005.

\bibitem{devries2017improved}
Terrance DeVries and Graham~W Taylor.
\newblock Improved regularization of convolutional neural networks with cutout.
\newblock {\em arXiv preprint arXiv:1708.04552}, 2017.

\bibitem{dong2020benchmarking}
Yinpeng Dong, Qi-An Fu, Xiao Yang, Tianyu Pang, Hang Su, Zihao Xiao, and Jun
  Zhu.
\newblock Benchmarking adversarial robustness on image classification.
\newblock In {\em Proceedings of the IEEE/CVF Conference on Computer Vision and
  Pattern Recognition}, pages 321--331, 2020.

\bibitem{dong2018boosting}
Yinpeng Dong, Fangzhou Liao, Tianyu Pang, Hang Su, Jun Zhu, Xiaolin Hu, and
  Jianguo Li.
\newblock Boosting adversarial attacks with momentum.
\newblock In {\em Proceedings of the IEEE conference on computer vision and
  pattern recognition}, pages 9185--9193, 2018.

\bibitem{hu2021naturalistic}
Yu-Chih-Tuan Hu, Bo-Han Kung, Daniel~Stanley Tan, Jun-Cheng Chen, Kai-Lung Hua,
  and Wen-Huang Cheng.
\newblock Naturalistic physical adversarial patch for object detectors.
\newblock In {\em Proceedings of the IEEE/CVF International Conference on
  Computer Vision (ICCV)}, 2021.

\bibitem{stochastic_depth}
Gao Huang, Yu Sun, Zhuang Liu, Daniel Sedra, and Kilian~Q Weinberger.
\newblock Deep networks with stochastic depth.
\newblock In {\em European conference on computer vision}, pages 646--661.
  Springer, 2016.

\bibitem{huang2021rpattack}
Hao Huang, Yongtao Wang, Zhaoyu Chen, Zhi Tang, Wenqiang Zhang, and Kai-Kuang
  Ma.
\newblock Rpattack: Refined patch attack on general object detectors.
\newblock In {\em 2021 IEEE International Conference on Multimedia and Expo
  (ICME)}, pages 1--6. IEEE, 2021.

\bibitem{glenn_jocher_2020_4154370}
Glenn Jocher, Alex Stoken, Jirka Borovec, NanoCode012, ChristopherSTAN, Liu
  Changyu, Laughing, tkianai, Adam Hogan, lorenzomammana, yxNONG, AlexWang1900,
  Laurentiu Diaconu, Marc, wanghaoyang0106, ml5ah, Doug, Francisco Ingham,
  Frederik, Guilhen, Hatovix, Jake Poznanski, Jiacong Fang, Lijun Yu,
  changyu98, Mingyu Wang, Naman Gupta, Osama Akhtar, PetrDvoracek, and Prashant
  Rai.
\newblock {ultralytics/yolov5}, Oct. 2020.

\bibitem{kingma2014adam}
Diederik~P Kingma and Jimmy Ba.
\newblock Adam: A method for stochastic optimization.
\newblock {\em arXiv preprint arXiv:1412.6980}, 2014.

\bibitem{li2018robust}
Yuezun Li, Daniel Tian, Ming-Ching Chang, Xiao Bian, and Siwei Lyu.
\newblock Robust adversarial perturbation on deep proposal-based models.
\newblock {\em arXiv preprint arXiv:1809.05962}, 2018.

\bibitem{liu2022efficient}
Siao Liu, Zhaoyu Chen, Wei Li, Jiwei Zhu, Jiafeng Wang, Wenqiang Zhang, and
  Zhongxue Gan.
\newblock Efficient universal shuffle attack for visual object tracking.
\newblock In {\em ICASSP 2022-2022 IEEE International Conference on Acoustics,
  Speech and Signal Processing (ICASSP)}, pages 2739--2743. IEEE, 2022.

\bibitem{liu2016ssd}
Wei Liu, Dragomir Anguelov, Dumitru Erhan, Christian Szegedy, Scott Reed,
  Cheng-Yang Fu, and Alexander~C Berg.
\newblock Ssd: Single shot multibox detector.
\newblock In {\em European conference on computer vision}, pages 21--37.
  Springer, 2016.

\bibitem{liu2021swin}
Ze Liu, Yutong Lin, Yue Cao, Han Hu, Yixuan Wei, Zheng Zhang, Stephen Lin, and
  Baining Guo.
\newblock Swin transformer: Hierarchical vision transformer using shifted
  windows.
\newblock In {\em Proceedings of the IEEE/CVF International Conference on
  Computer Vision}, pages 10012--10022, 2021.

\bibitem{madry2017towards}
Aleksander Madry, Aleksandar Makelov, Ludwig Schmidt, Dimitris Tsipras, and
  Adrian Vladu.
\newblock Towards deep learning models resistant to adversarial attacks.
\newblock {\em arXiv preprint arXiv:1706.06083}, 2017.

\bibitem{mohri2018foundations}
Mehryar Mohri, Afshin Rostamizadeh, and Ameet Talwalkar.
\newblock {\em Foundations of machine learning}.
\newblock MIT press, 2018.

\bibitem{papernot2017practical}
Nicolas Papernot, Patrick McDaniel, Ian Goodfellow, Somesh Jha, Z~Berkay Celik,
  and Ananthram Swami.
\newblock Practical black-box attacks against machine learning.
\newblock In {\em Proceedings of the 2017 ACM on Asia conference on computer
  and communications security}, pages 506--519, 2017.

\bibitem{redmon2017yolo9000}
Joseph Redmon and Ali Farhadi.
\newblock Yolo9000: better, faster, stronger.
\newblock In {\em Proceedings of the IEEE conference on computer vision and
  pattern recognition}, pages 7263--7271, 2017.

\bibitem{redmon2018yolov3}
Joseph Redmon and Ali Farhadi.
\newblock Yolov3: An incremental improvement.
\newblock {\em arXiv preprint arXiv:1804.02767}, 2018.

\bibitem{ren2015faster}
Shaoqing Ren, Kaiming He, Ross Girshick, and Jian Sun.
\newblock Faster r-cnn: Towards real-time object detection with region proposal
  networks.
\newblock {\em Advances in neural information processing systems}, 28, 2015.

\bibitem{srivastava2014dropout}
Nitish Srivastava, Geoffrey Hinton, Alex Krizhevsky, Ilya Sutskever, and Ruslan
  Salakhutdinov.
\newblock Dropout: a simple way to prevent neural networks from overfitting.
\newblock {\em The journal of machine learning research}, 15(1):1929--1958,
  2014.

\bibitem{szegedy2014intriguing}
Christian Szegedy, Wojciech Zaremba, Ilya Sutskever, Joan Bruna, Dumitru Erhan,
  Ian Goodfellow, and Rob Fergus.
\newblock Intriguing properties of neural networks.
\newblock In {\em 2nd International Conference on Learning Representations,
  ICLR 2014}, 2014.

\bibitem{thys2019fooling}
Simen Thys, Wiebe Van~Ranst, and Toon Goedem{\'e}.
\newblock Fooling automated surveillance cameras: adversarial patches to attack
  person detection.
\newblock In {\em Proceedings of the IEEE/CVF conference on computer vision and
  pattern recognition workshops}, pages 0--0, 2019.

\bibitem{vapnik1999nature}
Vladimir Vapnik.
\newblock {\em The nature of statistical learning theory}.
\newblock Springer science \& business media, 1999.

\bibitem{wei2019transferable}
Xingxing Wei, Siyuan Liang, Ning Chen, and Xiaochun Cao.
\newblock Transferable adversarial attacks for image and video object
  detection.
\newblock In {\em Proceedings of the 28th International Joint Conference on
  Artificial Intelligence}, pages 954--960, 2019.

\bibitem{wu2020dpattack}
Shudeng Wu, Tao Dai, and Shu-Tao Xia.
\newblock Dpattack: Diffused patch attacks against universal object detection.
\newblock {\em arXiv preprint arXiv:2010.11679}, 2020.

\bibitem{wu2020making}
Zuxuan Wu, Ser-Nam Lim, Larry~S Davis, and Tom Goldstein.
\newblock Making an invisibility cloak: Real world adversarial attacks on
  object detectors.
\newblock In {\em European Conference on Computer Vision}, pages 1--17.
  Springer, 2020.

\bibitem{xie2017adversarial}
Cihang Xie, Jianyu Wang, Zhishuai Zhang, Yuyin Zhou, Lingxi Xie, and Alan
  Yuille.
\newblock Adversarial examples for semantic segmentation and object detection.
\newblock In {\em Proceedings of the IEEE international conference on computer
  vision}, pages 1369--1378, 2017.

\bibitem{yamada2018shakedrop}
Yoshihiro Yamada, Masakazu Iwamura, and Koichi Kise.
\newblock Shakedrop regularization.
\newblock 2018.

\bibitem{ying2019overview}
Xue Ying.
\newblock An overview of overfitting and its solutions.
\newblock In {\em Journal of physics: Conference series}, volume 1168, page
  022022. IOP Publishing, 2019.

\bibitem{zhang2017mixup}
Hongyi Zhang, Moustapha Cisse, Yann~N Dauphin, and David Lopez-Paz.
\newblock mixup: Beyond empirical risk minimization.
\newblock {\em arXiv preprint arXiv:1710.09412}, 2017.

\bibitem{zhao2020object}
Yusheng Zhao, Huanqian Yan, and Xingxing Wei.
\newblock Object hider: Adversarial patch attack against object detectors.
\newblock {\em arXiv preprint arXiv:2010.14974}, 2020.

\bibitem{zhong2020random}
Zhun Zhong, Liang Zheng, Guoliang Kang, Shaozi Li, and Yi Yang.
\newblock Random erasing data augmentation.
\newblock In {\em Proceedings of the AAAI conference on artificial
  intelligence}, volume~34, pages 13001--13008, 2020.

\end{thebibliography}
}

\end{document}


\title{T-SEA: Transfer-based Self-Ensemble Attack on Object Detection \\ (Supplementary Material)}

\author{First Author\\
Institution1\\
Institution1 address\\
{\tt\small firstauthor@i1.org}
\and
Second Author\\
Institution2\\
First line of institution2 address\\
{\tt\small secondauthor@i2.org}
}
\maketitle

\section{Details of Constrained Data Augmentation}

\section{Additional Ablation Study}

\subsection{Ablation Study on Patch Scale}

\subsection{Ablation Study on Patch Cutout Ratio}

\subsection{Ablation Study on Shakedrop Type}

\subsection{Ablation Study on TV Loss}

\section{Visualization}

\subsection{Activation of Feature Map}

\subsection{Cross Dataset Attack}

\subsection{Patch Scheduler Adjustment}

\section{Detailed Experimental Results}

\subsection{T-SEA with Different Base Attack Method}

\subsection{T-SEA across Different Datasets}

\begin{table*}[h]
\setlength\tabcolsep{2.5pt}
\centering
\small
\begin{tabular}{@{}cc|c|ccccccc|c@{}}
\toprule
\multicolumn{2}{c|}{\multirow{2}{*}{{\textbf{Method}}}} & White Box $\downarrow$ & \multicolumn{7}{c|}{Black Box $\downarrow$} & \multirow{2}{*}{\textbf{Black-Box Avg$\downarrow$}} \\ \cmidrule(r){3-10}
& & YOLO v5 & YOLO v2 & YOLO v3 & YOLO v3tiny & YOLO v4 & YOLO v4tiny & Faster R-CNN & SSD &  \\ \midrule

\multirow{3}{*}{Adam} & AdvPatch & \textbf{13.39} & 46.90 & 54.93 & 29.20         & 62.16 & 46.15 & 47.71 & 50.73 & \textbf{48.25} \\
                    & E-baseline & \textbf{20.21}   & 32.31 & 45.89 & 40.82         & 59.36 & 53.09 & 42.85 & 48.52 & \textbf{46.12}  \\
                     & T-SEA(ours) & \textbf{1.37}    & 11.61 & 16.77 & 10.73         & 32.53 & 17.55  & 17.14 & 30.78 & \textbf{19.59}  \\ \midrule
\multirow{3}{*}{SGD} & AdvPatch & \textbf{20.39} & 45.29 & 64.97 & 40.84         & 71.48 & 55.43 & 55.01 & 58.75 & \textbf{55.97}  \\
                    & E-baseline & \textbf{3.38}   & 18.52 & 26.36 & 25.56         & 31.66 & 30.44 & 24.21 & 36.33 & \textbf{27.58}  \\
                     & T-SEA(ours)  & \textbf{1.66}   & 8.66  & 11.93 & 9.19          & 22.22 & 13.75 & 15.23 & 29.95 & \textbf{15.85}  \\ \midrule
\multirow{3}{*}{MIM}  & AdvPatch  & \textbf{11.91} & 30.60 & 40.88 & 18.02         & 49.24 & 17.86 & 31.89 & 46.12 & \textbf{33.51}  \\
                    & E-baseline & \textbf{2.41}  & 14.63 & 24.63 & 15.60         & 30.70 & 25.45  & 24.28 & 34.85 & \textbf{24.31}  \\
                     & T-SEA(ours)   & \textbf{1.43}  & 8.64  & 11.44 & 7.90          & 20.62 & 14.34 & 15.35 & 29.75 & \textbf{15.43}  \\ \midrule
\multirow{3}{*}{BIM}  & AdvPatch  & \textbf{8.47} & 18.91 & 32.94 & 29.66         & 44.07 & 34.26 & 29.09 & 41.60 & \textbf{32.93}  \\
                    & E-baseline  & \textbf{2.05} & 15.52 & 20.78 & 19.13 & 26.66 & 25.08  & 18.80 & 33.50 & \textbf{22.78}  \\
                     & T-SEA(ours)  & \textbf{1.62}  & 10.62 & 15.17 & 12.80         & 28.78 & 17.79  & 16.85 & 32.05 & \textbf{19.15}  \\ \midrule
\multirow{3}{*}{PGD}  & AdvPatch & \textbf{13.58} & 31.24 & 38.31 & 30.56         & 46.77 & 38.25 & 33.75 & 52.04 & \textbf{38.70}  \\
                    & E-baseline  & \textbf{2.17}   & 16.81 & 20.59 & 16.17 & 20.97 & 21.53 & 19.70 & 35.51 & \textbf{21.61}  \\
                     & T-SEA(ours)  & \textbf{1.60}   & 10.70 & 15.11 & 13.40 & 18.36 & 17.47 & 18.34 & 33.06 & \textbf{18.06}  \\ \midrule
\end{tabular}
\caption{The Results of T-SEA with Different Base Attack Method. Here we select five classical base attack methods to verify the method robustness of T-SEA, which includes optimization-based methods (e.g., Adam and SGD) and iterative methods (e.g., MIM, BIM and PGD). The results shows that the proposed T-SEA can enhance the performance of all these methods, demonstrating its good adaptability to different methods.}
\label{exp:base_attack_methods}
\end{table*}

\clearpage

{\small
\bibliographystyle{ieee_fullname}
\bibliography{egbib}
}